# Federated Learning for Cross-block Oil-water Layer Identification

Bingyang Chen[a], Xingjie Zeng[a], Weishan Zhang[a,*], Zhitao Li[b], Zhaoxiang Hou[a], Siyuan Xu[c], Chenyu Sun[a], Dakuang Han[d]

[a]*School of Computer Science and Technology, China University of Petroleum, Qingdao, 266580, China.*

[b]*National University Science Park, China University of Petroleum Qingdao, 266580, China.*

[c]*PetroChina Dagang Oilfield Company, Tianjin, 300280, China*

[d]*PetroChina Research Institute of Petroleum Exploration & Development, Beijing, 100083, China.*

**Abstract:** Cross-block oil-water layer(OWL) identification is essential for petroleum development. Traditional methods are greatly affected by subjective factors due to depending mainly on the human experience. AI-based methods have promoted the development of OWL identification. However, because of the significant geological differences across blocks and the severe long-tailed distribution(class imbalanced), the identification effects of existing artificial intelligence(AI) models are limited. In this paper, we address this limitation by proposing a dynamic fusion-based federated learning(FL) for OWL identification. To overcome geological differences, we propose a dynamic weighted strategy to fuse models and train a general OWL identification model. In addition, an F1 score-based re-weighting scheme is designed and a novel loss function is derived theoretically to solve the data long-tailed problem. Further, a geological knowledge-based mask-attention mechanism is proposed to enhance model feature extraction. To our best knowledge, this is the first work to identify OWL using FL. We evaluate the proposed approach with an actual well logging dataset from the oil field and a public 3W dataset. Experimental results demonstrate that our approach significantly out-performs other AI methods.

## 1. INTRODUCTION

OWL identification facilitates the efficiency and success of oil exploration and development[1]. Conventional methods usually utilize well logging data for lithology evaluation, reservoir parameter prediction, and thus OWL identification. The main methods, cross-plotting [2] and statistical analysis [3]. However, the method based on expert experience combined with mathematical statistics is seriously affected by human subjectivity, prone to error accumulation and high cost. The large logging data generated by the extensive deployment of wells presents opportunities and challenges for further formation understanding[4]. The multi-solution and uncertainty of well logging data make OWL identification increasingly difficult, thus AI technology is urgently required to improve efficiency and interpretation compliance[5].

Traditional machine learning methods have been successfully applied to identify OWL and lithology within a block, including Artificial neural network (ANN) [6-7], Support vector machine (SVM) [8-9], Random Forest (RF) [10-11], XGBoost [12-13]. However, machine learning has limited feature extraction capability and is prone to overfitting, thus neural networks show great potential for OWL identification. Due to the sequence property of the logging curve, OWL identification is a sequence classification problem along the formation's longitudinal dimension. Long Short-Term Memory(LSTM) networks are employed to extract curve features for lithology classification[14]. Considering the feature extraction capability and computational efficiency, Bi-directional Gate Recurrent Unit (Bi-GRU) is applied to further improve the model effect[15]. A convolutional neural network (CNN) [16] is used for OWL identification due to its excellent feature extraction capabilities. There are two model training approaches in existing studies: multi-well cross-validation and adjacent well training. The cross-validation method utilizes multiple wells to test another well (within-block) and repeats the process until all wells have been tested once. However, in a practical scenario, we need to train the model with limited data (e.g. one well) to identify the OWL of most remaining wells. The adjacent well training identifies the target wells using the same block adjacent wells (well distance within 400m). However, when significantly geological differences exist between wells, resulting in weak generalization of the training model to identify OWL. In addition, the data long-tailed distribution makes it difficult for the model to learn the features of small sample categories (e.g., oil layers) [17].

To address the above challenges, we train a general model as a pre-training model, and then finetune the model using any well in a new block to identify OWL of other wells(within about 2km of the finetuned well). Because of the risk of data transfer leakage when fusing data to train a general model, an FL [18] is employed to learn the geological features of each block distributively, and model fusion instead of data fusion is employed to preserve data security. Therefore, a dynamic fusion FL approach is proposed to improve model effectiveness and generalization. In addition, an effective re-weighting scheme is designed to improve the long-tailed problem. Further, a geological knowledge-based mask attention mechanism is designed to improve model feature extraction capability. This paper focuses on well logging data and integrates mud logging, perforating and production monitoring data for OWL identification.

∗ Corresponding author: Weishan Zhang, School of Computer Science and Technology, China University of Petroleum, Qingdao, 266580, China.

E-mail addresses: bychen.@s.upc.edu.cn (Bingyang Chen), xj.zeng@mail.utoronto.ca (Xingjie Zeng), zhangws@upc.edu.cn (Weishan Zhang), hxg@s.upc.edu.cn (Zhaoxiang Hou), xusiyuan_dg@163.com (Siyuan Xu)

Our primary contributions of the paper are as follows:

(1) A novel AI method for cross-block OWL identification. To the best of our knowledge, this is the first work to study OWL identification based on FL.

(2) A FL-based dynamic weighted fusion strategy to address the geological differences across blocks and data security problems.

(3) An effective re-weighting scheme to address the long-tailed problem. The effectiveness of the loss is verified by derivation and experiments, and the class balance loss [19] is combined to further improve the model performance.

(4) A geological knowledge-based mask attention mechanism. As we show through experimental evaluations, performs better than that of the original Transformer.

The remainder of the paper is organized as follows. Section 2 discussed related work. Section 3 presents the approach for OWL identification in detail. Section 4 evaluates the approach. Section 5 concludes the paper and points out the future work.

## 2. Related Work

### 2.1 Federated learning framework

FL [18] is a distributed learning framework consisting of multiple clients and a central server, which enables federated modeling while ensuring data security. Each client trains a model using local data and uploads it to the central server for average fusion. Then the central server builds a global model and dispatches it to each client for the next iteration[20]. Since the data is stored locally, it effectively avoids data leakage through model fusion. Horizontal FL is a method to federally train samples that are in different clients but have the same features to improve model generalization，which has been widely used in industry[21] and medical[22] fields. However, the increase of clients inevitably reduces the model training speed[23]. Therefore, we can learn the model fusion method[24] using model upload time as a weight to improve model communication efficiency.

### 2.2 Transformer for feature extraction

Transformer [25] is originally a model consisting of six encoders and six decoders for machine translation. Encoder aims to convert the input sequence into a vector, and the Transformer mentioned later refers to its encoder. Each encoder (more details in section 3.3) contains a self-attention layer and a feed forward neural network layer. The attention score in self attention layer is calculated as follows:

$$Attention(Q, K, V) = softmax(QK^T/\sqrt{d_k}) \quad (1)$$

Where $Attention(Q, K, V)$ is the attention score of a current word and a target word in a sentence, $Q$ is the current word query vector，$K, V$ are a set of target word key-value vectors. The purpose of softmax is to normalize the scores of all words and $\sqrt{d_k}$ is to reduce the K-dimension's effects on attention scores.

Skip connect adding between two sublayers aims to prevent gradient disappearance, layer normalization adding aims to reduce the numerical differences between sample features. The Transformer utilizes a self attention mechanism for fast parallel computation and it does rely on the past hidden states to capture the previous words. Thus it possesses a stronger ability to extract intrinsic features. However, self-attention achieves weak performance in extracting local features. Inspired by Fan [26], the attention effects can be further improved when combined with practical scenarios.

### 2.3 Long-tailed problem

Long-tailed distributions are common in practical classification or recognition tasks, since recognition results are likely to be biased towards dominant classes (head) and perform weakly in small sample classes (tail) [27]. There are two methods to address the long-tailed distribution problem: re-sampling and re-weighting. The former reverse weights the sampling frequency of sample per class according to sample size [28], the latter changes the loss of classification. Since some tasks (e.g., target detection in the image domain) are too complex to sample, most studies[29-30] use the re-weighting approach. Class balanced loss (CB) [19] introduces a concept of effective number, representing the expected number of samples per class to rebalance the loss. However, the effective number may be zero in the practical scenario, so the idea of CB is worth learning and can be further improved.

## 3. Approach

In this section, we elaborate on the proposed FL-based approach for cross-block OWL identification. We first describe how a dynamic weighted fusion strategy is applied to FL. Then we present a geological knowledge-based mask attention mechanism for OWL identification. Finally, we design and theoretically prove the proposed re-weighting scheme.

### 3.1 Problem statement

Considering the significant differences across blocks, a GFL[31] framework is introduced to enable models to learn from blocks' geological features. As shown in Fig.1, each client in an FL training round, corresponding to one block, trains a model with local data and sends it to a central server after model filtering. Then the central server builds a global model using a dynamic weighted fusion and dispatches it to each client for the next iteration. In this way, the block data leakage is eliminated since only model transmission but without data communication. In general, this study makes the following assumptions:

(1) There are multiple blocks as clients in the FL. Due to the significant geological differences across blocks, the model trained by each client individually has weak generalization.

(2) All OWL identification tasks are five classification tasks on clients and have the identical label space.

(3) The OWL identification model is shared from the server to each client (block).

(4) The local data of each block are never transmitted to preserve data security.

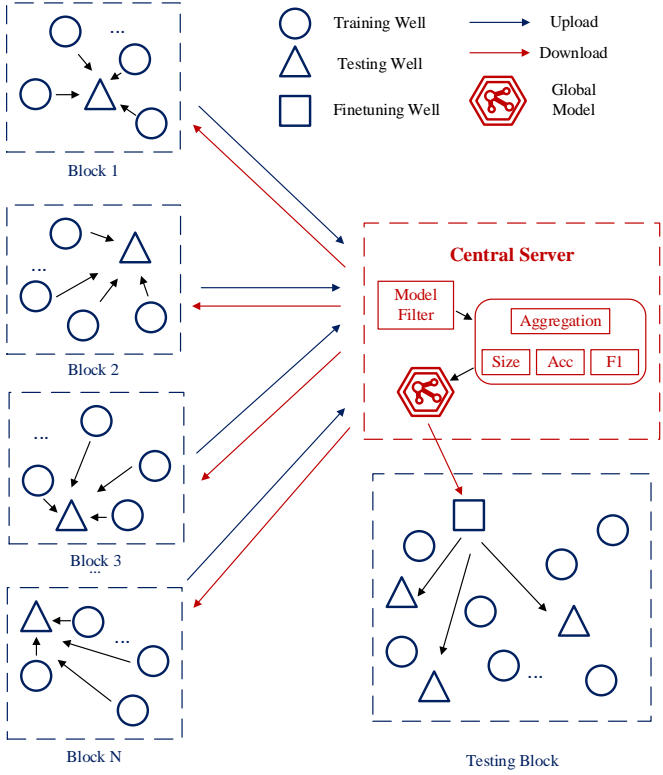

Fig. 1. Dynamic weighted fusion-based FL framework for OWL identification

*3.2 Federated learning with dynamic fusion*

FL enables the model to learn the dominant features of each client distributively to compensate for the original deficiencies of the local model, which eliminates local data sharing to preserve information security. We design a dynamic weighted fusion strategy based on original FL to improve the model effects. The proposed dynamic fusion process is shown in Algorithm 1, which can be divided into 6 steps.

Step1:Each block (client) downloads parameters from the central server for initialization and trains the model using local data.

Step2:The F1 Score is employed to rebalance the loss and combined with the smoothed CB [20] function to address the long-tail problem. (Line4-8)

Step3: In the model filtering, only those with higher F1 scores than the previous version are uploaded to the central server. (line9-11)

Step4: In the model fusion, the accuracy, F1 scores and data size of each block are used as weights to update the global model. (Line17-20)

Step5:The parameters of the global model are dispatched to each block for the next training.

Step6:Return to the first step until model convergence.

| Algorithm 1: Dynamic Fusion-based FL for OWL identification |
|---|
| **Input:** The private data of each block $b^i$ |
| **Output:** Our oil-water layer identification model |
| 1 /* Each Client (Oil-bearing block) $b^i$ */ |
| 2  Download $\theta, F_i^{last}$ from server |
| 3  $Acc_i, F_i \rightarrow train\left(f_\theta(x_i^b)\right)$ |
| 4  $F_g = 2y^b f_\theta(x_i^b)/\left(y_i^b + f_\theta(x_i^b)\right)$ |
| 5  F_Loss $= -\left(y_i^b(y_i^b - 0.5F_g)\log f_\theta(x_i^b) + (1 - y_i^b) \cdot 0.5F_g\log\left(1 - f_\theta(x_i^b)\right)\right)$ |
| 6  $CB = (1 - \beta)/(1 - \beta^{s+1})$ |
| 7  $L_{D^{tr}}(\theta) \leftarrow \frac{1}{D^{tr}}\sum_{(x_i^b, y_i^b) \in D^{tr}} CB \cdot F\_Loss(f_\theta(x_i^b), y_i^b)$ |
| 8  $\theta \leftarrow \theta - \gamma\nabla_\theta L_{D^{tr}}\left(f_\theta(x_i^b)\right)$ |
| 9  if $F_i > F_i^{last}$: |
| 10    $F_i^{last} = F_i$ |
| 11    Send $f_\theta(x_i^b), F_i^{last}, Acc_i, Size_i$ to server |
| 12 |
| 13 /*Central Server*/ |
| 14 Initialize global model parameters $\theta$ |
| 15 $F_i^{last}, Acc_i, Size_i \leftarrow$ Receive() |
| 16 Uploaded Client Model $f_\theta(x_i^b) \leftarrow$ ReceiveModel() |
| 17 **for** each round $t=1,2,3…,N$ **do** |
| 18    $w_i \leftarrow softmax(F_i^{last}, Acc_i, Size_i)$ |
| 19    Global Model $F_\theta(x) \leftarrow Aggregate\left(f_\theta(x_i^b), w_i\right)$ |
| 20 **end for** |
| 21 Dispatch(Global Model) |

**Model Filtering**: Each block $b^i(i = 1,2,..,n)$, namely, a client downloads the global model parameters $\theta$ and the F1 score $F_i^{last}$ of the previous version from the server. One well in a block is randomly selected as the test set $D^{te}$, and the other wells are used as the training set $D^{tr}, (x_i^b, y_i^b) \in D^{tr}$. The client's training data is $x_i^b$, its corresponding label is $y_i^b$, and the client model is $f_\theta(x_i^b)$. The Accuracy $Acc_i$ and F1 score $F_i$ of the current round are obtained by model testing. We assign F1 score gradient information $F_g$ and then design a novel loss function $F\_Loss$ to rebalance loss(the derivation process is in Section 3.4) due to data long-tailed distribution. In addition, we combine the smoothed $CB$ to build a loss $L_{D^{tr}}(\theta)$. The parameter $\theta$ update process using gradient descent as follows:

$$\theta \leftarrow \theta - \gamma\nabla_\theta L_{D^{tr}}\left(f_\theta(x_i^b)\right) \quad (2)$$

where $\gamma$ is a learning rate and $\nabla_\theta$ is a derivative of the parameter $\theta$. The F1 score in current round $F_i$ is used for model filtering. Assign $F_i$ to $F_i^{last}$ if current F1 score greater than the previous version $F_i^{last}$. Then the client model $f_\theta(x_i^b)$, F1 score $F_i^{last}$, accuracy $Acc_i$ and data size $Size_i$ are uploaded to the central server.

evaluation weights(Eq.6) and fuse the models to update the global model $F_\theta(x)$ (Eq.7).

$$w_i \leftarrow softmax(F_i^{last}, Acc_i, Size_i) \quad (6)$$
$$F_\theta(x) \leftarrow Aggregate(f_\theta(x_i^b), w_i) \quad (7)$$

where $w_i$ is the fusion weight of the ith client model, $Aggregate()$ is the model fusion function, and $f_\theta(x_i^b), w_i$ are parameters.

The flow chart of the proposed dynamic weighted fusion-based FL for OWL identification is shown in Fig.2. First, the central server initializes the model and dispatches it to each client. Each client then trains the local model and only uploads models when F1 scores greater than the previous version. The central server dynamically updates the global model with model's accuracy, F1 score, and data size. One well from the target block is employed to finetune the global model if maximun rounds reached, and other wells in the block are randomly selected for testing.

### 3.3 Logging curves feature extraction with mask attention

Feature extraction is a critical step and directly affects OWL identification effect. Therefore, we incorporate domain knowledge to improve Transformer, and the model structure is shown in Figure 3.

The specific layer(class label) corresponding to a certain depth data $x_i (i = 1,2,....k)$ is affected by the nearby data, namely, the label $y_i$ is associated with $x_{i-j}, x_i, x_{i+j} (i-j \geq 1 \& i+j \leq k)$. Therefore, the d-dimensional features are selected as the model input and sliding by stratigraphic longitudinal dimension with a window of size $d*k$. A $d*k$ matrix is the training data, and a specific layer $y_{(k+1)/2}$ is the class label to train the model. In the meandering river sedimentation, sandstone and mudstone are compacted with each other, which makes reservoir and non-reservoir alternately appear. There is usually only a single class of reservoir (e.g., water layer) between non-reservoirs. Therefore, we propose a mask attention mechanism with this domain knowledge to improve the Transformer. Specifically, the data class labelled as non-reservoir is masked instead of directly removing the non-reservoir data. Calculating attention after adding mud well logging features, thus improving the efficiency of OWL identification.

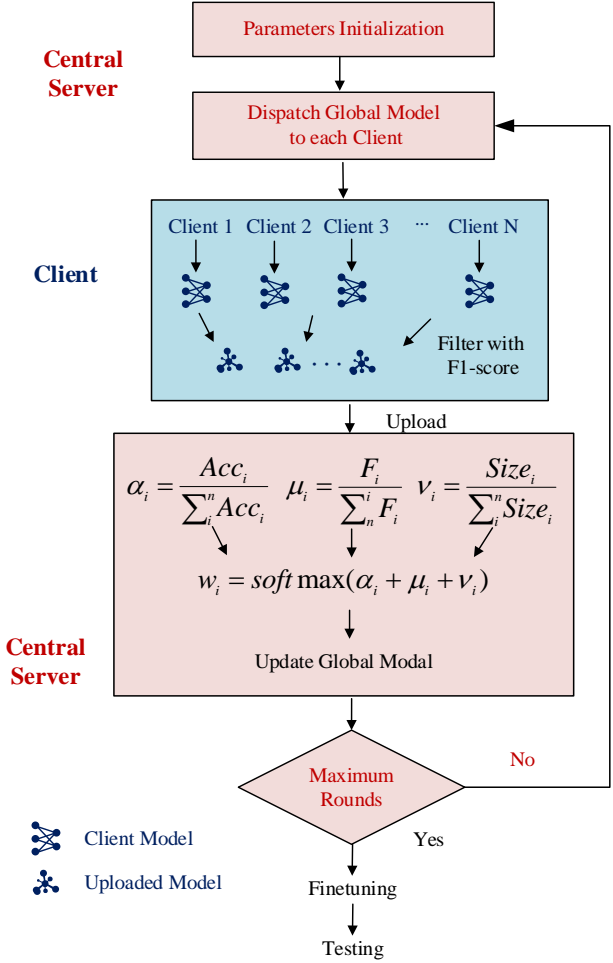

Fig. 2. Flow chart of the proposed approach for OWL identification

**Dynamic fusion:** In each round (*t=1,2,3...,N*) fusion process, the calculation of each client's evaluation weight $\alpha_i$, $\mu_i$ and $\nu_i$ are as follows:

$$\alpha_i = Acc_i / \sum_n^i Acc_i \quad (3)$$
$$\mu_i = F_i^{last} / \sum_n^i F_i^{last} \quad (4)$$
$$\nu_i = Size_i / \sum_n^i Size_i \quad (5)$$

where $Acc_i$ is the accuracy of the ith client, similarly, $F_i^{last}, Size_i$ is the corresponding F1 scores and data size. We then use the softmax function to normalize the

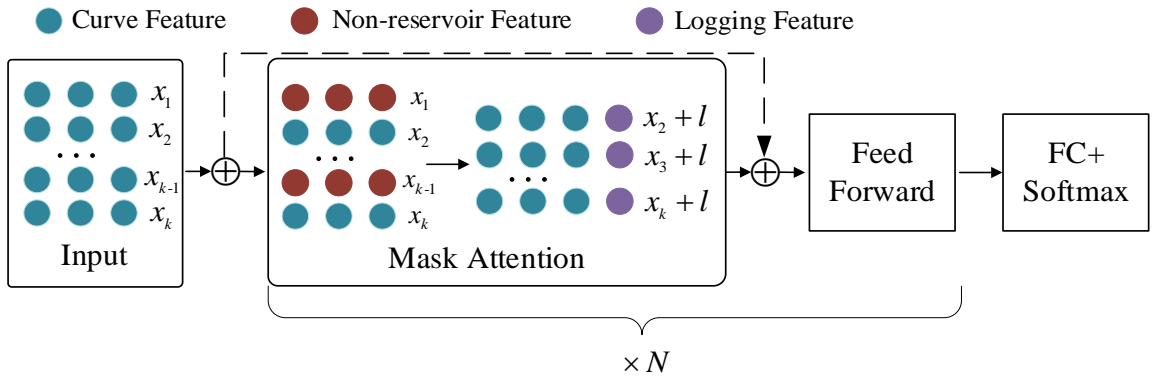

*Fig. 3. Proposed Transformer with mask attention for oil-water layer identification*

In client training, the data size of each batch is $Batch$, and its dimension $[Batch * d * k]$ is transformed into the specified dimensional to input model, which is sent to a Feed Forward Network after mask attention calculation. This is one layer of Transformer, through $N$ layers to extract features. $\oplus$ denotes skip connect and layer normalization, the former overcomes the limitation of gradient disappearance and weight matrix degradation, the latter normalize multiple curve features of the same depth. The output of the Transformer is then fed into the Fully Connected Network for mapping to higher dimensions. The final dimension is reduced to the specified neurons number, which also represents the classification confidence. Then the classification probability is calculated by softmax. The loss function proposed in Section 3.4 is employed for back-propagation to train the model.

*3.4 Mathematical formulation of the proposed loss function*

There is some inverse relationship between accuracy and loss, thus we attempt to take the opposite of accuracy as a loss. However, the original accuracy has no gradient, and we need to build gradients for it. F1 score is similar to accuracy.

**Reasoning 1**：Building gradients for accuracy：

For simplicity, the accuracy of the two-classification task is taken as an example, and multi-classification tasks can be migrated. Suppose $x$ is the input, $y \in \{0,1\}$ is the target, and $f_\theta(x) \in [0,1]$ is the model. The predicted class is $M$, the target class is $N$, $M_1, N_1$ denote the one hot vector with 1 at $m, n$ position respectively. The inner product (Eq.8) is one when the predicted category is consistent with the target category, and zero when it is inconsistent. The accuracy $Acc$ (Eq.9) can be defined as:

$$M_1 \cdot N_1^T = \begin{cases} \begin{bmatrix}0\\1\end{bmatrix} \cdot [0 \ 1] = 1 (m = n) \\ \begin{bmatrix}0\\1\end{bmatrix} \cdot [1 \ 0] = 0 (m \neq n) \end{cases} \quad (8)$$

$$Acc = \left(M_1(x) \cdot N_1^T(x)\right) \quad (9)$$

The output of the model should be a probability distribution to ensure its derivability, so we replace $M_1(x)$ with a probability distribution as follows:

$$Acc = f_\theta(x) N_1^T(x) = f_\theta(x) y - (1 - f_\theta(x))(1 - y) \quad (10)$$

**Definition 1**：It is better to transform $f_\theta(x)$ to $\log f_\theta(x)$ for loss calculation.

Proof: take the simplest activation function sigmoid $\sigma$ as an example, which is calculated as follows:

$$f_\theta(x) = \sigma(g_\theta(x)) \quad (11)$$

The derivative as:

$$\sigma'(g_\theta(x)) = 1/(1 + e^{-g_\theta(x)}) \quad (12)$$

Suppose $y = 1$, then:

$$Acc = -f_\theta(x) = -\sigma(g_\theta(x)) \quad (13)$$

The gradient is calculated as follows:

$$-\nabla_\theta f_\theta(x) = -f_\theta(x)(1 - f_\theta(x))\nabla_\theta g_\theta(x) \quad (14)$$

Our assumption is that $y = 1$, so a larger gradient when $f_\theta(x) = 0$ and a smaller gradient when $f_\theta(x) = 1$ is what we expect. However, $f_\theta(x)(1 - f_\theta(x))$ gets the minimum value when $f_\theta(x)$ takes 0.5, when zero or one takes the maximum value instead(Eq.14), which does not match our expectation.

The cross-entropy loss function (Eq.15) and its gradient (Eq.16) are calculated as follows：

$$CE = -y \log f_\theta(x) - (1 - y) \log(1 - f_\theta(x)) \quad (15)$$

$$-\nabla_\theta \log f_\theta(x) = -(1 - f_\theta(x))\nabla_\theta g_\theta(x) \quad (16)$$

Since $f_\theta(x)$ is removed in the calculation of the gradient, the gradient becomes larger when $f_\theta(x) = 0$ and smaller when $f_\theta(x) = 1$, which is consistent with our expectation. Comparison of Eq.14 and Eq.16 indicates $-\nabla_\theta f_\theta(x)$ is inferior to $-\nabla_\theta \log f_\theta(x)$, so we can obtain a definition as follows:

$$f_\theta(x) \to \log f_\theta(x) \quad (17)$$

**Reasoning 2**：Rebalance the loss with F1 Score to obtain a novel loss function: $F\_Loss = -\left(y(y - 0.5F_g)\log f_\theta(x) + (1 - y) \cdot 0.5F_g \cdot \log(1 - f_\theta(x))\right)$

Due to the long-tailed distribution, F1 score is employed to rebalance the loss function. Similar to the accuracy, we first build gradients for F1 score as follows

$$F_g = 2y f_\theta(x)/(y + f_\theta(x)) \quad (18)$$

The gradient descent update process is as follows:

$$\begin{aligned}
-\nabla_\theta & \frac{2y f_\theta(x)}{(y + f_\theta(x))} \\
&= -2 \frac{\nabla_\theta y f_\theta(x) \cdot (y + f_\theta(x))}{(y + f_\theta(x))^2} + 2 \frac{y f_\theta(x) \cdot \nabla_\theta f_\theta(x)}{(y + f_\theta(x))^2} \\
&= -2\nabla_\theta f_\theta(x) \frac{y(y + f_\theta(x)) - y f_\theta(x)}{(y + f_\theta(x))^2} \\
&= -2\nabla_\theta f_\theta(x) \frac{y(y + f_\theta(x)) - 0.5F_g(y + f_\theta(x))}{(y + f_\theta(x))^2} \\
&= -\nabla_\theta f_\theta(x)(y - 0.5F_g) \frac{2}{(y + f_\theta(x))}
\end{aligned} \quad (19)$$

Taking $2/(y + f_\theta(x))$ as a scaling factor, we only focus on the sample gradient as follows:

$$-(y - 0.5F_g) \cdot \nabla_\theta f_\theta(x) \qquad (20)$$

The Eq.20 is transformed according to **Definition 1** as follows:

$$-\left(y(y - 0.5F_g)\nabla_\theta log f_\theta(x) + (1-y) \cdot 0.5F_g \cdot \nabla_\theta log(1 - f_\theta(x))\right) \qquad (21)$$

Thus the improved loss function $F\_Loss$ as follows:

$$F\_Loss = -\left(y(y - 0.5F_g)log f_\theta(x) + (1-y) \cdot 0.5F_g \cdot log(1 - f_\theta(x))\right) \qquad (22)$$

We balance the positive sample cross-entropy using $y - 0.5F_g$ and the negative sample cross-entropy using $0.5\ F_g$. To further address the long-tailed problem, we also smoothed the CB[20] to avoid effective samples number is zero(Eq.23). The final loss function (Eq. 24) is obtained as follows:

$$CB = (1 - \beta)/(1 - \beta^{s+1}) \qquad (23)$$

$$L_{D^{tr}}(\theta) \leftarrow \frac{1}{D^{tr}}\sum_{(x_i^b, y_i^b) \in D^{tr}} CB \cdot F\_Loss(f_\theta(x_i^b), y_i^b) \qquad (24)$$

where $\beta$ is the hyperparameter with a value of 0.9999, $s$ is the effective number of samples per class, and $x_i^b, y_i^b$ is the training data and label of the ith block.

## 4. Evaluation

In this section, we evaluate the model performance on an actual well logging dataset and a 3w public dataset[32]. First, we explore the impact of model fusion approaches on FL, as well as the network structure of the client model. Second, we analyse the effects of the model optimization approach and the model parameters in ablation experiments.

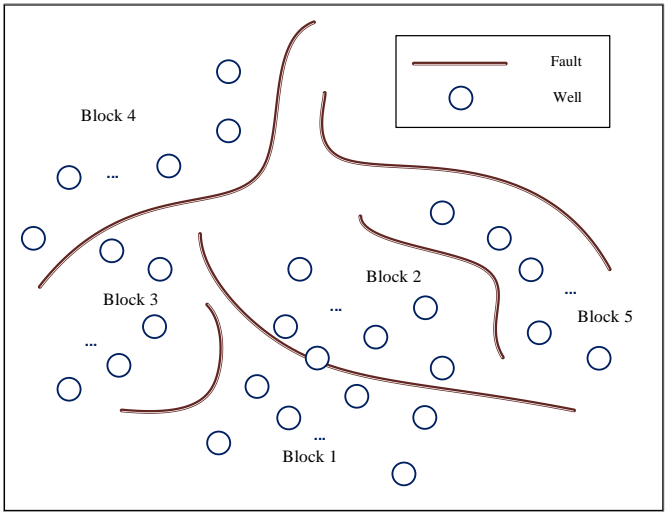

Fig. 4. Schematic diagram of well location distribution

*4.1 Dataset description*

We take five blocks in an oil field as an example to identify cross-block OWL, where significant geological differences exists across blocks due to fault separation. The study region is meandering river sedimentation. A schematic diagram of the well distribution is shown in Fig. 4. Five curve features (SP, CAL, AC, RA25, DEPTH) commonly possessed among the blocks are selected for the study. Mud logging, perforating and production monitoring data are utilized to obtain realistic labels for improving training effects. The logging curves are sampled at 0.125m depth intervals, and 49 sampling points are used as one sample in the experiment (more details in section 4.3.2). One well in each block is randomly selected and the distribution of layer classes is shown in Table 1.

**Table 1. Statistics of well logging dataset**

| Class | B1 | B2 | B3 | B4 | B5 |
|---|---|---|---|---|---|
| Oil layer | 5 | 71 | 22 | 28 | 3 |
| Dry layer | 203 | 31 | 38 | 88 | 8 |
| Water layer | 736 | 338 | 55 | 90 | 380 |
| Oily water layer | 66 | 0 | 122 | 240 | 15 |
| Oil and water layer | 31 | 69 | 951 | 808 | 0 |

Where B1 means Block 1, others are similar.

The dataset(3W) of oil well anomaly detection(WAD) is used as a public dataset to demonstrate the effectiveness and generalization of our approach. On the one hand, differences across wells can be analogous to geological differences across blocks. On the other hand, since it is numerical data collected from the sensor, the 3W dataset is similar to well logging data on both source and type. Since the data size of 3W is large, some real instances(Table 2) in 3W are selected in the comparison experiment. The data contains seven features ('timestamp','P-PDG','P-TPT','T-TPT','P-MON-CKP','T-JUS-CKP','P-JUS-CKGL ') and four classes.

**Table 2. Statistics of 3w dataset**

| Class | Well 1 | Well 2 | Well 3 | Well 5 | Well 14 |
|---|---|---|---|---|---|
| 0 | 5 | 10 | 1 | 3 | 0 |
| 2 | 0 | 1 | 2 | 0 | 0 |
| 3 | 1 | 0 | 0 | 0 | 3 |
| 4 | 2 | 7 | 0 | 3 | 2 |

**TABLE 3 EXPERIMENT ENVIRONMENT**

| | GPU | RAM | Python | CUDA |
|---|---|---|---|---|
| Server | RTX3090 | 24G | 3.8.10 | 11.2 |
| Client 1 | RTX 3090 | 24G | 3.8.10 | 11.2 |
| Client 2 | RTX 3090 | 24G | 3.8.10 | 11.2 |
| Client 3 | RTX 3080 | 10G | 3.8.10 | 11.2 |
| Client 4 | GTX1080Ti | 11G | 3.8.10 | 11.2 |

*4.2 Comparative experiment*

*4.2.1 Comparisons of model fusion in federated learning*

FL effects of different fusion methods are explored on both datasets, the experimental environments are shown in Table3. The layers of client model is five, learning rate is 0.0001, ReLU is activation function and AdamW is optimizer. The iteration rounds of client and server is 10,30 respectively.The model is evaluated with accuracy, F1 score, and training speed. Four blocks(Block 1,2,3,5) data are selected for training client model, which then uploaded to the central server for aggregation. One well in Block 4 is randomly selected to finetune the global model, and then another well in that block is selected for testing each time. Finally, the global model effect is evaluated by testing the average value. Similarly, four wells(well 2,5,11,14) in 3W dataset are selected as clients data, 80 percent of which are used for training and the rest are used for testing. Then the trained model are uploaded to the central server for updating the global model. Five percent data of well 1 are randomly selected for finetuning the global model each time, and the rest are used for testing. The global model is also evaluated with test average.

**Table 4. THE EFFECTS OF DIFFERENT MODEL FUSION METHOD**

|  | Oil-water layers identification | | | Anomaly detection | | |
| --- | --- | --- | --- | --- | --- | --- |
| Model | Accuracy(%) | F1(%) | Time | Accuracy (%) | F1(%) | Time |
| FedAvg | 70.84 | 39.92 | 12min15s | 83.79 | 56.81 | 6h59min38s |
| FedF1 | 77.24 | **63.95** | 13min07s | 86.72 | 78.90 | 6h50min14s |
| Fed(Acc+F1+Size) | 88.96 | 78.28 | 12min40s | 90.18 | 86.81 | 6h57min05s |
| FilterAvg | 71.36 | 42.61 | 8min22s | 84.51 | 59.98 | 5h48min25s |
| FilterF1 | 83.27 | 61.24 | 9min17s | 92.11 | 78.93 | 5h54min06s |
| Filter(Acc+F1+Size) | 90.11 | 81.74 | 8min45s | 94.85 | 90.33 | 6h01min13s |

FedAvg、FedF1、Fed(Acc+F1+Size) mean all client models are uploaded. Where FedAvg means average fusion, FedF1 means fused by F1 score. Fed(Acc+F1+Size) means fused by accuracy, F1 score and data size. FilterAvg、FilterF1、Filter(Acc+F1+Size) mean client model is uploaded when its F1 score higher than the previous version. The fusion methods correspond to those described above.

OWL identification results show the model filtering does not greatly improve the performance of the models, but the training speed is significantly improved. FedF1 performs much better than FedAvg, especially the F1 score, which illustrates the deficiency of average fusion. Interestingly, the F1 score of FedF1 is slightly higher than that of FilterF1 in the OWL experiment, indicating that part of the effective client model is filtered out during model filtering. For example, in the model uploading, the F1 score of Block 3 model (e.g., 60%) is much higher than that of Block 5 (e.g., 30%). But it is filtered out only because it is slightly smaller than the previous version (e.g., 61%), which makes the optimal global model is not available. Filter(Acc+F1+Size) performs better than Fed(Acc+F1+Size), indicating that the fusion weight of F1 score is weakened in the fusion process and the optimal solution is obtained. The experimental results on both datasets further demonstrate that the effectiveness of our approach.

*4.2.2 Model effects with finetune rounds*

In this section, the global model effects with different finetune rounds are further discussed. One well in Block 4 is randomly selected to finetune with 20 rounds, and another three wells for testing. Fig.5 shows that almost all wells converge in the 16[th] round, indicating the generalization of the proposed model. The higher performance of well B indicates that is geological features are similar to those of the finetune wells.

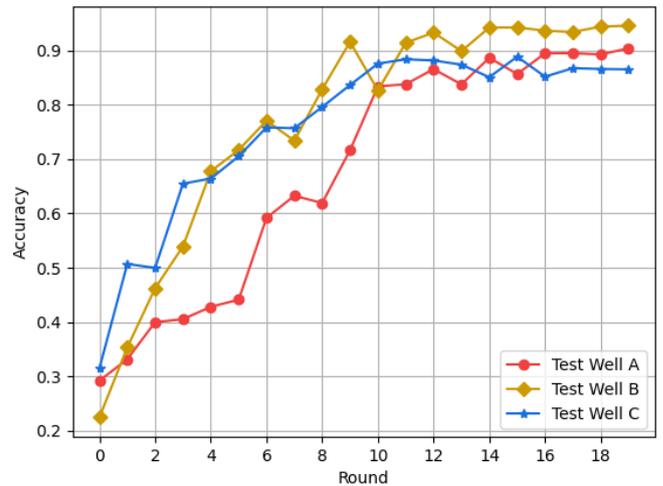

(a) Accuracy with finetune rounds

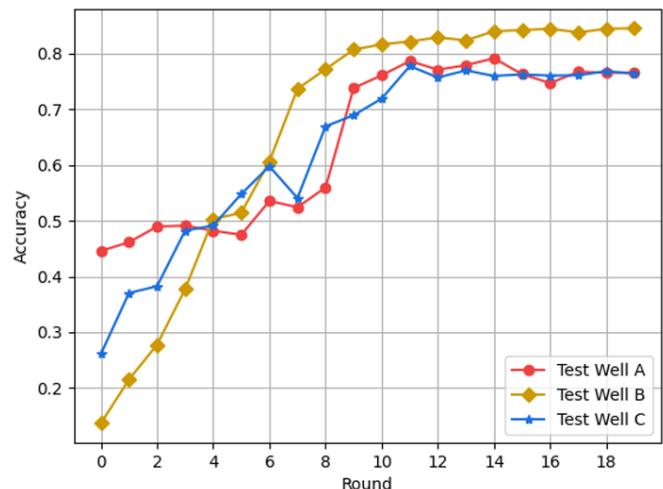

(b) F1 score with finetune rounds

Fig. 5. Model performance with finetune rounds

### 4.2.3 Comparison with some baselines

The parameters of baselines are similar to Transformer in client model comparison experiment. In within-block OWL experiment, one well in Block 4 for training and it adjacent well for testing. And in cross-block OWL experiment, three wells in Block 3 for training and one well in Block 4 for testing. Similar processing in 3W dataset to demonstrate the effectiveness of our approach. In within-well experiment, 80 percent data of a well for training and the rest for testing. In cross-well experiment, well 2 data is employed for training model to test well 1.

As shown in Table 6, the performance of within-block is generally higher than that of cross-block in well logging experiment, indicating that models unable to learn stratigraphic features adequately because of significant geological differences. Machine learning models perform better than deep learning in within-block experiment, indicating within-block features are similar and easy to learn while deep learning models require a larger amount of data. Because of a strong adaptability, Transformer performs best in cross-block experiment. Similar conclusions can conclude in 3W dataset. Furthermore, both machine learning models and Transformer achieve better performance in cross-well experiment, illustrating the differences of cross-well are smaller than that of cross-block. Excellent performance of Transformer on both datasets makes it is the main model architecture of clients. Compared with using only Transformer for classification, FL(section 4.2.1) significantly improves model effect and generalization.

**Table 6. THE PERFORMANCE OF DIFFERENT MODAL ON BOTH DATASETS**

| Model | Within-block | Cross-block | Within-well | Cross-well |
|---|---|---|---|---|
| Random Forest | 92.38/ **91.12** | 41.52/ 13.75 | 97.57/ **98.18** | 78.15/ 32.04 |
| XGBoost | 93.32/ **92.98** | 62.68/ 23.25 | 98.62/ **98.37** | 81.50/ 39.51 |
| MLP | 65.05/ 25.63 | 57.16/ 23.67 | 61.29/ 26.60 | 48.52/ 29.88 |
| CNN | 70.08/ 30.51 | 69.39/ 27.31 | 51.95/ 33.81 | 42.82/ 18.52 |
| Bi-LSTM | 74.56/ 23.65 | 70.08/ 24.54 | 46.35/ 34.11 | 48.67/ 22.51 |
| Transformer | 93.17/ 74.77 | 67.24/ 37.08 | 94.35/ 88.58 | 83.35/ 39.37 |

The values(percent) in the table represent the Accuracy and F1 Score, respectively.

### 4.3 Ablation experiment

#### 4.3.1 Effects of different model optimization methods

In this section, we further explore the impact of different optimization approaches on the each model effect improvement. In cross-block experiment, three wells of four blocks(Block 1,2,3,5) are randomly selected for training model to test one will in Block 2. In cross-well experiment, similarly, well 2 is selected for training model to test well 1. In Table 7, *Feature Adding* indicates the model effect with only logging features added, *Mask Attention* indicates the model effect only optimized by masking with non-reservoir features. *CB-CE*, *CB-F* indicate the model effect of combing *CB* in the cross-entropy and *F_Loss* function respectively. Final optimized model(FOM) represents model training effect after integrating all optimization methods. Since the WAD experiment does not involve *Feature Adding* and *Mask Attention*, FOM effect is the same as that of *CB-F*.

Due to the long-tailed problem of data, the model is mainly evaluated by F1 score, aided by accuracy. OWL identification experiment results show that each block's model effects is improved adding log features, which is also in accordance with the geological knowledge. The model effect is also improved by mask attention, especially the accuracy, which is slightly stronger than that of *Feature Adding*. The use of non-reservoirs for OWL identification, rather than simply discarding non-reservoirs information, indicating non-reservoir features contribute to model improvement. Experiments show that the proposed loss function significantly improves the effectiveness of the model, especially the F1 score. Compared with *CB-CE*, *CB-F* further improves the F1 score of the models, indicating the effectiveness of the proposed re-weighting scheme. In the WAD experiment, *CB-CE* only has a slight increase in F1 score, since the long-tailed problem of 3W data was not as serious as the well logging.

**Table 7. THE EFFECTS OF DIFFERENT MODEL OPTIMIZATION**

| Model | B1 | B2 | B3 | B5 | Well |
|---|---|---|---|---|---|
| Transformer | 69.85/ 22.27 | 78.38/ 21.08 | 67.24/ 37.08 | 50.88/ 17.22 | 83.35/ 39.37 |
| Feature Adding | 78.04/ 32.92 | 75.61/ 31.48 | 81.91/ 30.81 | 72.04/ 27.91 | -- |
| Mask Attention | 79.17/ 31.48 | 83.15/ 30.67 | 86.68/ 42.83 | 75.92/ 27.54 | -- |
| CB-CE | 74.04/ 38.91 | 75.54/ 41.38 | 74.37/ 42.72 | 70.04/ 38.28 | 80.88/ 42.17 |
| CB-F | 77.87/ 43.09 | 75.85/ 47.32 | 78.81/ 50.11 | 74.59/ 43.96 | 83.75/ 53.85 |
| FOM | 83.06/ 52.35 | 84.43/ 49.68 | 83.67/ 56.31 | 75.85/ 47.73 | 83.75/ 53.85 |

Where B1 means Block1, others are similar. The values(percent) in the table represent the Accuracy and F1 Score, respectively.

FOM performs better in all experiments, further demonstrating the effectiveness of the proposed optimization method. In addition, results show that Block 3 model generally performs better, while Block 5 models perform weaker. This is since Block 3 is close to Block 5, which is consistent with the domain knowledge that distance between wells is positively correlated with geological differences.

#### 4.3.2 Parameters analysis

The optimal size of the sliding window ($d * k$) is explored in this section, where we focus on the optimal value of the coverage distance $k$. Due to the inconsistent layer thickness，information near the sampling point will be extracted insufficiently if k is small, while several layers will be covered if k is large. Therefore, the setting of k value has a great impact on the recognition effect. The one-side coverage distance (from $(k+1)/2$ to $k$) of the sampling point is set as 1-6m in the experiments. Three wells in Block 4 are employed for testing. The original Transformer is used for exploring the optimal value. F1 score is used as the main evaluation indicator in this process. Table 8 shows that although Well C performs best at 5m, the difference between its effect and that at 3m is very small. In addition, the other two wells performs best at 3m. Therefore, 3m is the optimal value of one-side coverage distance.

**Table 8. THE EFFECTS OF DIFFERENT SLIDE DISTANCE**

| Distance | Well A | Well B | Well C |
|---|---|---|---|
| 6m | 78.38/27.95 | 73.19/33.37 | 66.67/25.04 |
| 5m | 65.31/31.40 | 73.11/32.68 | 57.25/**27.48** |
| 4m | 68.60/35.07 | 74.22/33.25 | 57.81/24.74 |
| 3m | 67.24/37.08 | 73.34/36.02 | 58.29/27.42 |
| 2m | 66.55/29.80 | 70.40/31.25 | 50.63/25.78 |
| 1m | 69.39/30.24 | 73.42/30.23 | 48.83/22.87 |

The values(percent) in the table represent the Accuracy and F1 Score respectively.

## 5. Conclusions and Future Work

This paper proposes a novel and effective approach for cross-block OWL identification based on dynamic fusion FL. First, we propose a FL approach based on dynamic weighted fusion for cross-block OWL identification. Second, we design an effective re-weighting scheme to address the long-tailed problem by rebalancing loss with F1 score. Third, we present a geological knowledge-based mask attention mechanism to enhance the effect of OWL identification. The results show that the proposed model achieves high performance in terms of accuracy, F1 score, and training speed.

In the future, we will further study the identification of water flooded layers to explore the dynamic changes of small layers during oilfield production and development.

Acknowledgements

The research is supported by the National Natural Science Foundation of China (62072469), National Key R&D Program (2018yfe0116700), Shandong Natural Science Foundation (ZR2019MF049), China University of Petroleum Graduate Student Innovation Project Funding Program (YCX2021126) and the China Scholarship Council.